\documentclass[10pt,twocolumn,letterpaper]{article}

\usepackage{iccv}              

\definecolor{iccvblue}{rgb}{0.21,0.49,0.74}
\usepackage[pagebackref,breaklinks,colorlinks,allcolors=iccvblue]{hyperref}

\usepackage{algorithm} 
\usepackage{algpseudocode} 
\usepackage{multirow}

\def\paperID{8193} 
\def\confName{ICCV}
\def\confYear{2025}

\title{Boosting Adversarial Transferability via Residual Perturbation Attack}

\author{Jinjia Peng$^{1,\dag}$ \quad
  Zeze Tao$^{1,\dag}$ \quad
  Huibing Wang$^{2,*}$ \quad
  Meng Wang$^{3}$ \quad
  Yang Wang$^{3,*}$
  \\
  $ ^{1} $School of Cyber Security and Computer, Hebei University\\
  $ ^{2} $College of Information and Science Technology, Dalian Maritime University\\
  $ ^{3} $School of Computer and Information Engineering, Hefei University of Technology\\
\texttt{\small  \{pengjinjia, zeze\}@hbu.edu.cn, huibing.wang@dlmu.edu.cn, eric.mengwang@gmail.com} \\
  \texttt{\small  yangwang@hfut.edu.cn}
}

\begin{document}
\maketitle

\begin{abstract}
Deep neural networks are susceptible to adversarial examples while suffering from incorrect predictions via imperceptible perturbations.
Transfer-based attacks create adversarial examples for surrogate models and transfer these examples to target models under black-box scenarios.
Recent studies reveal that adversarial examples in flat loss landscapes exhibit superior transferability to alleviate overfitting on surrogate models.
However, the prior arts overlook the influence of perturbation directions, resulting in limited transferability.
In this paper, we propose a novel attack method, named Residual Perturbation Attack (ResPA), relying on the residual gradient as the perturbation direction to guide the adversarial examples toward the flat regions of the loss function.
Specifically, ResPA conducts an exponential moving average on the input gradients to obtain the first moment as the reference gradient, which encompasses the direction of historical gradients. Instead of heavily relying on the local flatness that stems from the current gradients as the perturbation direction,
ResPA further considers the residual between the current gradient and the reference gradient to capture the changes in the global perturbation direction.
The experimental results demonstrate the better transferability of ResPA than the existing typical transfer-based attack methods, while the transferability can be further improved by combining ResPA with the current input transformation methods. The code is available at 
\url{https://github.com/ZezeTao/ResPA}.
\end{abstract}

{
  \renewcommand{\thefootnote}{\fnsymbol{footnote}}
  \footnotetext[0]{\dag Equal contribution.\quad * Corresponding authors.} 
}

\section{Introduction}
\label{sec:intro}

\begin{figure}[!t]
\centering
\includegraphics[width=0.47\textwidth]{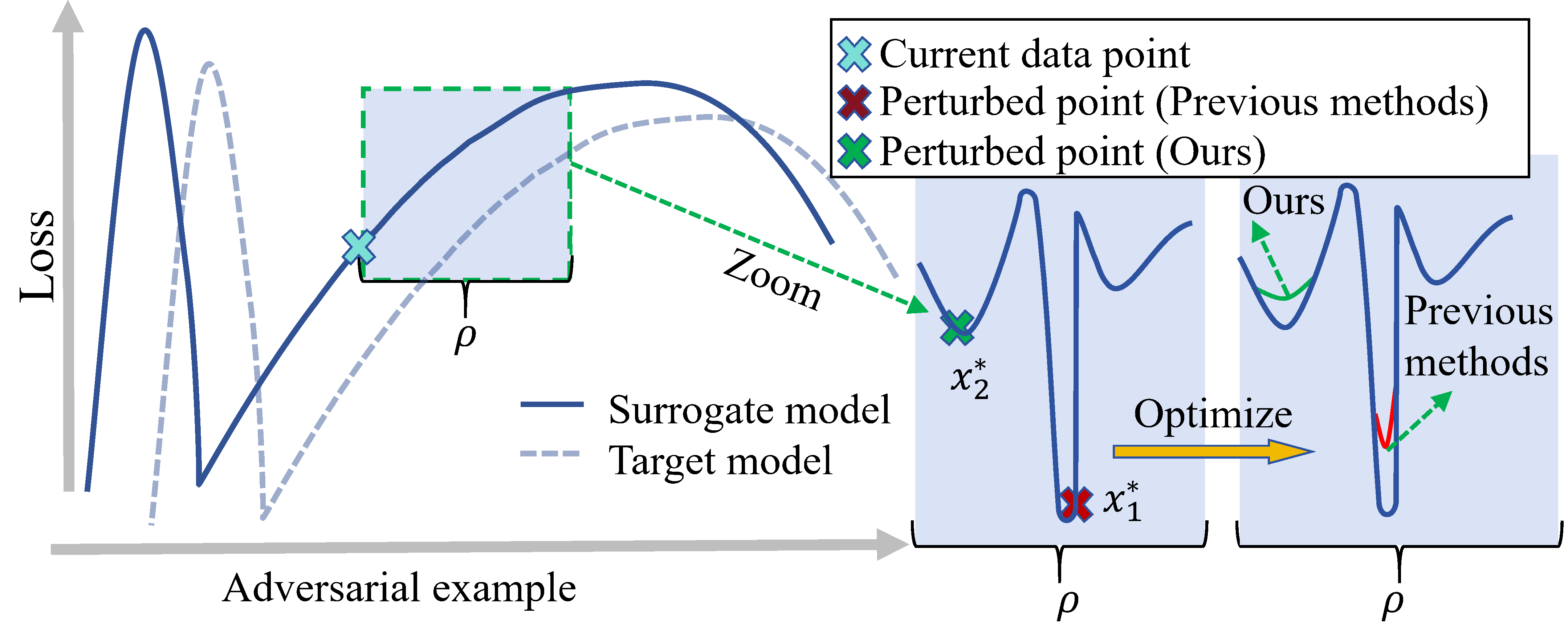}
\caption{Comparison between ResPA and previous methods in searching for the perturbed point. $\rho$ is the perturbation radius. Within $\rho$, extremely sharp regions often exist. Previous methods flatten the local region by optimizing the loss of the perturbed point in the sharpest areas. In contrast, ResPA optimizes the loss of the perturbed point, which is beneficial for the global situation.}
\label{fig:raodong}
\end{figure}
Deep neural networks (DNNs) have exhibited outstanding capabilities in various language and vision processing applications.
 However, adversarial examples \cite{FGSM,50IFGSM,51CVPR2018mifgsm} have been demonstrated to be indiscernible from natural ones, but can mislead a model to generate incorrect predictions.
 Additionally, adversarial examples generated from the surrogate model can also transfer to other target models.
 The transferability of adversarial examples renders adversarial attacks viable in real-world scenarios \cite{selfdviv,wang2022sci,ijcai2020,wang2020tip, qi2025unsupervised}.



Based on the adversary's level of knowledge about the target model, adversarial attacks can be classified into white-box attacks \cite{FGSM,50IFGSM,randomIFGSM} and black-box attacks \cite{51CVPR2018mifgsm, 34CVPR2021vmifgsm, 2024BSR}. In the white-box scenario, adversaries possess comprehensive knowledge of the target models, encompassing their structures, parameter weights, and the training loss function. In contrast, in the black-box scenario, attackers create adversarial examples using a white-box surrogate model, which are subsequently transferred to the black-box target model.
In real-world applications, DNN models are often hidden for users. Therefore, black-box attacks are generally more feasible. However, the over-parameterized DNNs with many sharp maxima are prone to trapping adversarial examples to be over-fitted on white-box surrogate models, resulting in poor adversarial transferability.

\begin{figure*}[!t]
\centering
\includegraphics[width=0.97\textwidth]{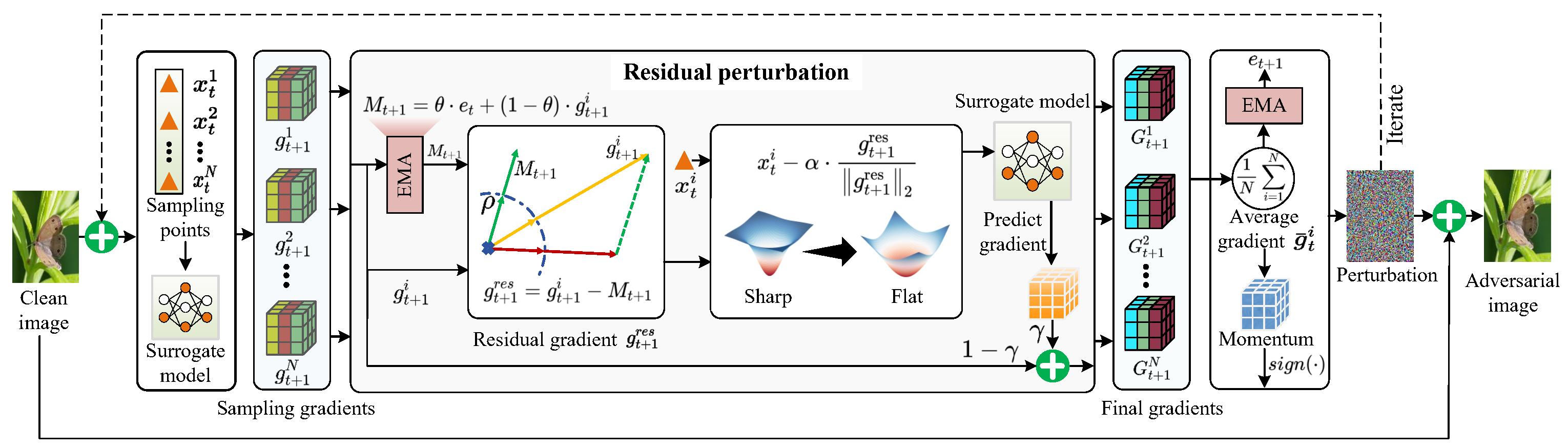}
\caption{An overview of the proposed ResPA attack. In the process of searching for flat regions, ResPA adopts the Exponential Moving Average to perform weighted averaging on the historical records of gradients and achieve the reference gradient. Then, ResPA generates the residual gradient as the perturbation direction, defined as the difference between the reference
gradient and the current gradient.}
\label{fig:proposed}
\end{figure*}
To alleviate the overfitting issue to enhance adversarial transferability, substantial techniques have been proposed, including gradient-based approaches \cite{51CVPR2018mifgsm, GRAadaptive, jifen2023gradient, anda2024strong}, input transformation-based approaches \cite{29attack2019NIsi,32attackICCV2021admix, tifgsm, 2024BSR, SSA2022ECCV}, and ensemble-based methods \cite{AdaEA_2023_ICCV,26attackensemble}.
In particular, recent flatness-based methods \cite{2024TPA,flat2022,PGN_2023_NIPS,wang2024pami,flat2023gnp} achieve state-of-the-art transferability performance by flatter maximum.
These studies can flatten the sharp regions of the loss landscape, thereby mitigating overfitting on the surrogate model and ultimately enhancing transferability of adversarial samples.
However, we observe that optimizing the loss of the perturbed point in excessively sharp regions with the perturbation radius may not enhance adversarial transferability. As shown in \cref{fig:raodong}, when excessively sharp regions exist with the perturbation radius, existing methods utilize the gradient as the perturbation direction to identify the perturbed point, which locates the perturbed point $x_1^*$ in the sharpest regions. Even after optimizing the flatness according to $x_1^*$, the point remains highly sharp, thereby failing to enhance the flatness of the entire loss surface effectively. Therefore, we turn to the perturbed point that deviates from the sharpest directions.

Based on the above, we propose a novel method, named Residual Perturbation Attack (ResPA), which adopts the residual gradient as the perturbation direction to search the perturbed point. Specifically, to prevent searching for the perturbed point in an overly sharp region, ResPA adopts the Exponential Moving Average (EMA) to perform weighted averaging on the historical gradients, thereby integrating the direction of the global gradients as the reference gradient.
To better capture the changes in the global perturbation direction, ResPA considers the residual gradient as the perturbation direction, defined as the difference between the current gradient and the reference gradient. In addition, it can capture the variations in the global perturbation direction, which could avoid excessive reliance on the flatness of the minimal perturbed point $x_1^*$. Instead of the perturbed point in the sharpest regions, ResPA identifies the perturbed point that are beneficial to the flatness of the entire loss surface. In summary, our contributions are summarized below:

\begin{itemize}

\item{To the best of our knowledge, we are the first to reveal that optimizing the loss of the perturbed point in overly sharp regions significantly hinders the transferability of flatness-based attack methods.}

\item{We propose a novel Residual Perturbation Attack method (ResPA), which employs the residual gradient as the perturbation direction to evaluate the flatness of local points. ResPA could capture the changes in the global perturbation direction, thereby avoiding searching for the perturbed point in the sharpest regions.}

\item{ResPA incorporates the proposed flatness term as a regularization, to maximize the loss function to flatten the loss surface.}
\end{itemize}
Experimental results demonstrate the better transferability of ResPA than the typical transfer-based attack methods. In addition, combining ResPA with the current input transformation method can further improve the transferability.

\section{Methodology}

\subsection{Preliminary}
Let $x$ represent the input image with $y$ as its corresponding true label. The classifier's output of the surrogate model\footnote{We adopt the varied deep models, e.g., Convolutional Neural Networks (CNNs), Vision Transformer (ViT) etc.} is denoted as $f^s(x)$ , and $J(x, y)=-\sum_{k=1}^{C} y_{k} \log f^s(x)$  is the cross-entropy loss function, where $C$ is the number of categories, while $y\in\{0,1\}^C$ is a one-hot encoded vector such that $y_k=1$ if $x$ belongs to the $k$-th class, and $y_k=0$ otherwise.
Given $x$, the goal of adversarial attacks is to identify an adversarial example $x^{adv}$ that deceives the classifier. Specifically, this adversarial example should prompt the classifier to output a label that differs from the true label, i.e., $f^s(x^{adv}) \neq f^s(x)$, while remaining imperceptible to human observers. This imperceptibility is ensured by following the constraint $\left\|x - x^{adv}\right\|_p < \epsilon$, where $\|\cdot\|_p$ denotes the $L_p$ norm  with $\epsilon > 0$ to define the perturbation magnitude. In this study, we concentrate on the $L_p$ norm for consistency with prior research.
Our goal is to maximize the subsequent optimization problem to generate the adversarial samples:
\begin{equation}
    \max _{x^{a d v}} J\left(x^{a d v}, y\right) \quad \text { s.t. } \left\|x - x^{a d v}\right\|_{\infty}<\epsilon.
    \label{eq:adv}
\end{equation}

Optimizing Eq. \eqref{eq:adv} requires calculating the gradient of the loss function, but this is not feasible in the black-box setting. Consequently, we generate the transferable adversarial examples on a surrogate model that can be used to attack other target models. The adversarial examples are encouraged to hoax target models to output wrong predictions, and the transferability can be quantitatively evaluated via the Attack Success Rate (ASR), calculated as follows:
\begin{equation}\label{eq:asr}
    A S R=\frac{1}{|\mathcal{X}|} \sum_{x \in \mathcal{X}} \mathbb{I}[f^t(x) \neq f^t(x^{adv})],
\end{equation}
where $\mathcal{X}$ denotes all the legitimate images. $f^t(\cdot)$ is the classifier's output of the target model. $\mathbb{I}(\cdot)$ is the indicator function, such that it equals to 1 once $f^t(x) \neq f^t(x^{adv})$, and 0 otherwise. 

\subsection{Residual Perturbation Attack}

At the core of our technique is to capture the changes in the global perturbation direction, while the over-reliance on the flatness of local points needs to be addressed. To this end, our proposed residual perturbation is elaborated upon in detail, followed by the process of searching flat maxima with our proposed residual perturbation.

\subsubsection{Residual Perturbation}

Previous methods \cite{2024TPA, flat2022, PGN_2023_NIPS, flat2023gnp} craft adversarial samples upon a locally flat region of the loss landscape, thereby elevating the adversarial transferability to unprecedented heights.
Within the given perturbation radius $\rho$, these methods measure \textbf{the flatness of the current point} by the difference between the loss of the perturbed point and the loss of the current point. However, in practical applications, there often exist excessively sharp regions within the perturbation radius, which weakens the effectiveness of the flatness term due to ``\textbf{excessively sharp regions}".
Since the existing methods \cite{2024TPA, flat2022, PGN_2023_NIPS, flat2023gnp} use the current gradient as the perturbation direction to search for the perturbed point, which is more likely to locate the perturbed point in the sharpest regions, as illustrated in \cref{fig:raodong}.
However, as can be seen from \cref{fig:raodong}, optimizing the loss of the perturbed point $x_1^*$ in the sharpest region does not effectively flatten the loss surface. Consequently, existing methods heavily rely on the flatness of the perturbed point in the local sharpest region, which is detrimental to the flatness of subsequent data points, thereby failing to achieve optimal transferability. To this end, we propose the Residual Perturbation Attack (ResPA) to optimize the loss of the perturbed point that are beneficial to global flatness, as shown in \cref{fig:proposed}. ResPA primarily consists of two components:
\begin{enumerate}
  \item Residual Perturbation: We define \textbf{a novel flatness term} using residual gradients, which addresses the ``\textbf{excessively sharp regions within the perturbation radius.}"
  \item Flat Maxima with Residual Perturbation: We incorporate \textbf{the proposed flatness term as a regularization term} into the maximization over the loss function to achieve a flat surface.
\end{enumerate}
In the $t$-th iteration, given the adversarial sample $x_t^{adv}$, the flatness of the loss function $\bar{J}$ at $x_t^{adv}$ is defined as:
\begin{equation}
\bar{J}\left(x_{t}^{adv},y\right)=\underbrace{\left[\min _{\|\delta\|_{1} \leq \rho} J\left(x_{t}^{adv }-\delta,y\right)-J\left(x_{t}^{adv},y\right)\right]}_{\text {flatness }},
    \label{eq:function}
\end{equation}
where $\delta$ is the perturbation vector with the same dimension as $x_{t}^{adv}$. Since it is difficult to track the exact minimal neighbor, the existing methods \cite{14sam,FGSM} utilize the gradient of the neighbor in the descent direction for iteration after two approximations:
\begin{equation}
    \min_{\|\delta\|_{1} \leq \rho}J(x_t^{adv}-\delta,y) \approx  J\left(x^{adv}_t-\rho \frac{d_t}{\left\|d_t\right\|},y\right),
    \label{eq:33}
\end{equation}
where $\rho$ denotes the radius to control the neighborhood size;
 $d_t$ denotes the gradient $\nabla_{x^{adv}_t} J(x^{adv}_t,y)$ at the current point, which determines the perturbation direction.
\cref{eq:33} encodes the loss of the perturbed point in the sharpest region. Since \cref{eq:33} relies solely on the current gradient direction, it fails to capture the variations in the global perturbation direction.
 As a result, it is not advantageous for subsequent data samples, which limits their transferability.


To address this limitation, we propose to exploit additional gradient information to search for the global perturbation direction. ResPA initially applies an exponential moving average (EMA) to the current gradient $\nabla_{x^{adv}_t} J(x^{adv}_t,y)$ to derive the first moment as the reference gradient. With increasing iterations, the reference gradient integrates the direction of all historical gradients, which is formulated as:
\begin{equation}
    M_{t+1} = \theta\cdot e_t+(1-\theta )\cdot \nabla_{x^{adv}_t} J(x^{adv}_t,y),
    \label{eq:lishi}
\end{equation}
where $\theta\in(0, 1)$ is the exponential decay factor; $e_t$ is calculated as the EMA of the previous average gradient according to \cref{eq:eee} and $e_0=0$.
To mitigate the excessive reliance on the perturbed point in the sharpest region, ResPA utilizes the residual gradient $g_{t+1}^{res}$ as the perturbation direction. This enables to capture the actual variations between the current and historical gradient direction. $g_{t+1}^{res}$ is formulated as follows:
\begin{equation}
     g_{t+1}^{res}=\nabla_{x^{adv}_t} J(x^{adv}_t,y) - M_{t+1}.
     \label{eq:cancha1}
\end{equation}

Finally, we utilize $g_{t+1}^{res}$ as the perturbation direction to locate the perturbed point, while \textbf{the proposed flatness term} can be obtained as follows:
\begin{equation}
   \bar{J}\left(x_{t}^{a d v},y\right) =  J\left(x_{t}^{a d v}-\rho \frac{g_{t}^{res}}{\left\|g_{t}^{res}\right\|},y\right)-J(x_t^{adv},y).
\end{equation}

\subsubsection{Flat Maxima with Residual Perturbation}

\textbf{Why can ResPA avoid searching for the perturbed point in excessively sharp regions?} In flat areas, the curvature of the loss surface is small, so that the current gradient $\nabla_{x^{adv}_t} J(x^{adv}_t,y)$ is relatively low.
In this case, as shown in \cref{eq:lishi}, the reference gradient $M_{t+1}$ is also small, and consequently, according to \cref{eq:cancha1}, the residual gradient $g_{t+1}^{res}$ is similarly small.
Therefore, in flat regions, neither previous methods nor our proposed ResPA will search for the perturbed point in overly sharp areas.


In sharp areas, the curvature of the loss surface is large, and the current gradient direction undergoes significant changes. Therefore, using the current gradient as the perturbation direction may be result in searching for the perturbed point in the steepest regions. In contrast, in ResPA, the residual gradient $g_{t+1}^{res}$ is defined as the difference between the current gradient $\nabla_{x^{adv}_t} J(x^{adv}_t,y)$ and the reference gradient $M_{t+1}$.
$M_{t+1}$ represents the average of historical gradients, which does not change significantly with abrupt variations in the current gradient. Therefore, when the current gradient $\nabla_{x^{adv}_t} J(x^{adv}_t,y)$ exhibits large fluctuations, the residual gradient $g_{t+1}^{res}$ can effectively suppress the current gradient to some extent, thereby avoiding searching for perturbation points in the sharpest regions.


In summary, when excessively sharp regions exist within the perturbation radius, ResPA avoids searching for the perturbed point in these overly sharp areas. Instead, it leverages the difference between the current gradient and historical gradients to identify the perturbed point that is more beneficial to the overall loss surface.


To incorporate the flatness term $\bar{J}\left(x_{t}^{ad v},y\right)$ into the optimization problem to improve the transferability of adversarial examples, ResPA employs \textbf{the flatness term as a regularization term} to impose constraints on the initial loss function. The primary goal is to jointly maximize the loss function and the flatness term. Following this strategy, ResPA could guide adversarial examples to flat regions.

Let $x_t^{adv}$ denote the input at the $t$-th iteration. $x_t^i = x_t^{adv}+\lambda _t^i$ is defined to be sampled within the neighborhood of $x_t^{adv}$, where $i = 1,2,\cdots,N$, such that $N$ represents the sample number. Here $\lambda_{t}^{i} \sim U\left[-(\beta \cdot \varepsilon)^{d}, (\beta \cdot \varepsilon)^{d}\right]$, where $U\left[a^{d}, b^{d}\right]$ stands for the uniform distribution in $d$ dimensions. The optimization problem of $J\left(x_{t}^{i}, y\right)$ is modified with the flatness term $\bar{J}\left(x_{t}^{a d v},y\right)$ in the $t$-th iteration as:
\begin{equation}
\begin{aligned}
\mathcal{L}\left(x_{t}^{i}, y\right) & =J\left(x_{t}^{i}, y\right)+ \gamma \cdot\bar{J}\left(x_{t}^{i}, y\right) \\
& =J\left(x_{t}^{i}, y\right)+\gamma \cdot\left[J\left(x_{t}^{*}, y\right)-J\left(x_{t}^{i}, y\right)\right] \\
& =(1-\gamma) \cdot J\left(x_{t}^{i}, y\right)+\gamma \cdot J\left(x_{t}^{*}, y\right),
\end{aligned}
\label{eq:optim}
\end{equation}
where $x_t^* = x_{t}^{i} - \rho \frac{g_{t}^{\text {res }}}{\left\|g_{t}^{\text {res }}\right\|}$ is the perturbed sample, $\gamma\in\left [ 0,1 \right ] $ is the penalty coefficient, and the regularization term is the flatness term, which can flatten the loss surface.


\begin{table*}[htbp]
\centering
\footnotesize
\begin{tabular}{|l|l|cccccccc|c|}
\hline
Model                    & Attack      & Inc-v3          & Res-50          & Vgg-19        & Den-121         & ViT           & Swin          &Inc-v3$_{ens3}$  &Inc-v3$_{ens4}$ & Average       \\ \hline \hline
\multirow{7}{*}{Inc-v3}  & MI \cite{51CVPR2018mifgsm}          & \textbf{100.0*} & 47.5            & 59.5          & 49.0            & 33.5          & 24.8          & 22.9          & 22.4          & 45.0          \\
                         & VMI \cite{34CVPR2021vmifgsm}        & \textbf{100.0*} & 65.5            & 70.1          & 66.9            & 44.6          & 40.4          & 39.1          & 39.1          & 58.2          \\
                         & GRA \cite{GRAadaptive}        & \textbf{100.0*} & 67.3            & 72.1          & 68.5            & 45.7          & 42.0          & 40.3          & 41.2          & 59.6          \\
                         & PGN \cite{PGN_2023_NIPS}        & \textbf{100.0*} & 73.4            & 76.8          & 74.5   & 52.7          & 44.8          & 42.8          & 43.5 & 63.6          \\
                         & AdaMSI \cite{adamsi}     & \textbf{100.0*} & 56.3            & 69.6          & 54.2            & 35.6          & 27.2          & 14.6          & 15.6          & 46.6          \\
                         & TPA \cite{2024TPA}        & 98.1*           & 68.2            & 70.7          & 68.2            & 46.8          & 43.3          & \textbf{44.6} & 42.2          & 60.3          \\
                         & ResPA (Ours) & \textbf{100.0*} & \textbf{75.7}   & \textbf{79.6} & \textbf{74.8}            & \textbf{54.0} & \textbf{45.8} & 42.3          & \textbf{44.3}          & \textbf{64.6} \\ \hline
\multirow{7}{*}{Res-50}  & MI \cite{51CVPR2018mifgsm}         & 65.8            & \textbf{100.0*} & 82.0          & 91.7            & 49.5          & 45.1          & 43.1          & 42.1          & 64.9          \\
                         & VMI \cite{34CVPR2021vmifgsm}        & 81.9            & 99.9*           & 91.6          & 96.1            & 67.0          & 63.5          & 65.2          & 64.5          & 78.7          \\
                         & GRA \cite{GRAadaptive}        & 87.0            & 99.9*           & 94.4          & \textbf{98.1}   & 72.8          & 68.3          & 72.7          & 70.3          & 82.9          \\
                        & PGN \cite{PGN_2023_NIPS}        & 88.1            & \textbf{100.0*} & 95.2          & 98.0            & 75.0          & 70.0          & 74.6          & 72.4 & 84.2          \\
                         & AdaMSI \cite{adamsi}     & 65.8            & \textbf{100.0*} & 89.4          & 91.2            & 48.2          & 43.6          & 36.6          & 36.8          & 64.0          \\
                         & TPA \cite{2024TPA}        & 85.2            & 98.7*           & 92.6          & 94.6            & 70.8          & 68.1          & 72.5          & 70.9          & 81.7          \\
                         & ResPA (Ours) & \textbf{88.8}   & \textbf{100.0*} & \textbf{95.5} & 98.0   & \textbf{75.5} & \textbf{71.3} & \textbf{75.2} &\textbf{72.6}          & \textbf{84.6} \\ \hline
\multirow{7}{*}{Den-121} & MI \cite{51CVPR2018mifgsm}         & 69.7            & 89.9            & 83.5          & \textbf{100.0*} & 55.1          & 51.4          & 49.1          & 49.5          & 68.5          \\
                         & VMI \cite{34CVPR2021vmifgsm}        & 84.5            & 96.2            & 91.3          & \textbf{100.0*} & 73.7          & 68.7          & 70.2          & 70.7          & 81.9          \\
                         & GRA \cite{GRAadaptive}        & 88.4            & 97.8            & 93.2          & \textbf{100.0*} & 78.6          & 75.3          & 78.9          & 76.0          & 86.0          \\
                        & PGN \cite{PGN_2023_NIPS}        & 89.5            & 97.5            & 94.9          & \textbf{100.0*} & 80.9          & \textbf{77.0}         & 80.0 & 77.7          & 87.2          \\
                         & AdaMSI \cite{adamsi}     & 76.2            & 93.8            & 92.7          & \textbf{100.0*} & 62.3          & 52.0          & 46.7          & 44.2          & 71.0          \\
                         & TPA \cite{2024TPA}        & 89.7            & 96.6            & 94.1          & 99.3*           & 79.4          & 75.3          & 79.3          & 76.8          & 86.3          \\
                         & ResPA (Ours) & \textbf{90.2}   & \textbf{97.9}   & \textbf{94.9} & \textbf{100.0*} & \textbf{82.1} & \textbf{77.0} & \textbf{80.3}          & \textbf{77.7} & \textbf{87.5} \\ \hline
\end{tabular}
\caption{The attack success rates (\%) on eight models by a single attack. The adversarial examples are generated on Inc-v3, Res-50,
 and Den-121 separately. Here * indicates the white-box model. The best results are bold. }
 \label{tab:black}
\end{table*}

We discuss more about \cref{eq:optim}, In particular:
\begin{enumerate}
  \item when $\gamma = 0$, the optimization problem is equivalent to solely maximizing the initial loss $J\left(x_{t}^{i}, y\right)$;
  \item when $\gamma = 1$, it aims at solely maximizing the perturbed point loss $J\left(x_{t}^{*}, y\right)$;
  \item when $\gamma \in (0,1)$, the optimization problem simultaneously balances both the initial loss and the perturbed point loss.
\end{enumerate}
The gradient of the current loss function can be formulated as follows:
\begin{equation}
    \nabla_{x_{t}^{i}} \mathcal{L}\left(x_{t}^{i}, y\right) \approx(1-\gamma) \cdot \nabla_{x_{t}^{i}} J\left(x_{t}^{i}, y\right)+\gamma \cdot \nabla_{x_{t}^{i}} J\left(x_{t}^{*}, y\right).
\end{equation}

Next, ResPA acquires the average gradient $\bar{g} _{t+1}$ over $N$ sampling points to be formulated as:
\begin{equation}
\bar{g}_{t+1}   = \frac{1}{N} \sum_{i=1}^{N}   \nabla_{x_{t}^{i}} \mathcal{L}\left(x_{t}^{i}, y \right),
\end{equation}
where $N$ represents the number of sampling points.
Next, we compute the EMA of the average gradient $\bar{g} _{t+1}$ to obtain the first-order moment $e_{t+1}$, which will be substituted into \cref{eq:lishi} in the next iteration to update the reference gradient $M_{t+1}$. The update rule for $e_{t + 1} $ is given by:
\begin{equation}
e_{t+1} = \theta\cdot e_t+(1-\theta )\cdot \bar{g} _{t+1},
\label{eq:eee}
\end{equation}
where $\theta$ is the exponential decay factor. Then the average gradient $\bar{g} _{t+1}$ is employed to update the momentum $g_{t+1}$, yielding:
\begin{equation}
    g_{t+1}=\mu \cdot g_{t}+\frac{\overline{g}_{t+1}}{\left\|\overline{g}_{t+1}\right\|_{1}},
\end{equation}
where $\mu$ is the decay factor. Finally, the adversarial samples $x_{t+1}^{a d v}$ are updated as follows:
\begin{equation}
    x_{t+1}^{a d v}=Clip_x^{\epsilon}(x_{t}^{a d v}+\alpha \cdot  \operatorname{sign}\left(g_{t+1}\right)),
\end{equation}
where $\operatorname{sign}(\cdot)$ is the sign function, $Clip_x^{\epsilon}(\cdot)$ denotes that the generated adversarial image is constrained within the $\epsilon$-ball neighborhood of the original image $x$, and $\alpha$ denotes the predetermined step size.






\section{Experiments}
\label{sec:exper}

\subsection{Experimental Setup}

\noindent\textbf{Dataset.}
We follow the convention of utilizing 1,000 origin images from the ILSVRC 2012 validation set \cite{russakovsky2015imagenet} to evaluate the performance of ResPA, mirroring the methodologies adopted in prior research \cite{34CVPR2021vmifgsm,32attackICCV2021admix}.
The models involved in this paper can classify these clean images with an accuracy of nearly 100\%. We also validate the effectiveness of the proposed method in the high-security-demand application scenario of person re-identification, with experiments conducted on Market-1501 \cite{10datasetmarket}.


%
%

\noindent\textbf{Models.}
We evaluate the attack success rate on 6 widely-used pre-trained models, namely Inception-v3 (Inc-v3) \cite{inception}, ResNet-50 (Res-50) \cite{resnet}, DenseNet-121 (Den-121) \cite{densenet}, VGGNet-19 (Vgg-19) \cite{vgg19}, Vision Transformer (ViT) \cite{vit}, and Swin Transformer (Swin) \cite{liu2021Swin} to validate the effectiveness of ResPA. Besides that, we consider adversarially trained models \cite{ensemblemodel}, specifically ens3-adv-Inception-v3 (Inc-v3$_{ens3}$) and ens4-adv-Inception-v3 (Inc-v4$_{ens4}$). Furthermore, seven state-of-the-art defense models are integrated, which exhibit exceptional robustness when defending against black-box attacks targeting the ImageNet dataset. The defense techniques encompass the high-level representation guided denoiser (HGD) \cite{HGD}, bit depth reduction (Bit-Red) \cite{bitred}, feature distillation (FD) \cite{FD2019feature}, JPEG compression (JPEG) \cite{JPEG}, neural representation purifier (NRP) \cite{NRP}, random resizing and padding (R\&P) \cite{RP2018mitigating}, as well as randomized smoothing (RS) \cite{RS2019certified}.

\noindent\textbf{Baseline Methods.}
In our experiments, six of the latest transfer-based attacks, namely MI \cite{51CVPR2018mifgsm}, VMI \cite{34CVPR2021vmifgsm}, GRA \cite{GRAadaptive}, PGN \cite{PGN_2023_NIPS}, AdaMSI \cite{adamsi}, and TPA \cite{2024TPA}, are taken into consideration. These attacks have exhibited superior performance in terms of success rates when benchmarked against earlier techniques such as FGSM \cite{FGSM} and I-FGSM \cite{50IFGSM}. Furthermore, we integrate the proposed ResPA with a variety of input transformations to affirm its effectiveness, such as DIM \cite{28attackDMICVPR2019}, TIM \cite{tifgsm}, SIM \cite{29attack2019NIsi}, Admix \cite{32attackICCV2021admix}, and SSA \cite{SSA2022ECCV}.

\noindent\textbf{Evaluation Metric.}
 In the experiment, we employ the attack success rate \cite{FGSM} in accordance with \cref{eq:asr} as the evaluation metric, which refers to the proportion of adversarial examples (among all generated ones) that can successfully mislead the target model.

\begin{table*}[!htbp]
\centering
\footnotesize
\begin{tabular}{|l|cccccccc|c|}
\hline
Attack     & Inc-v3        & Res-50*        & Vgg-19        & Den-121       & ViT           & Swin          & Inc-v3$_{ens3}$    & Inc-v3$_{ens4}$    & Average       \\ \hline \hline
DIM \cite{28attackDMICVPR2019}       & 85.9          & \textbf{100.0} & 93.2          & 96.9          & 67.3          & 61.8          & 67.9          & 63.4          & 79.6          \\
DIM+Ours   & \textbf{94.3} & \textbf{100.0} & \textbf{97.4} & \textbf{98.7} & \textbf{87.6} & \textbf{75.9} & \textbf{88.1} & \textbf{85.9} & \textbf{91.0} \\ \hline
TIM  \cite{tifgsm}      & 71.1          & \textbf{100.0} & 84.5          & 92.7          & 60.5          & 49.4          & 55.3          & 51.5          & 70.6          \\
TIM+Ours   & \textbf{94.8} & \textbf{100.0} & \textbf{96.8} & \textbf{98.9} & \textbf{87.8} & \textbf{74.2} & \textbf{87.7} & \textbf{87.2} & \textbf{90.9} \\ \hline
SIM \cite{29attack2019NIsi}       & 84.2          & \textbf{100.0} & 89.7          & 97.7          & 63.4          & 57.2          & 65.5          & 60.8          & 77.3          \\
SIM+Ours   & \textbf{92.4} & \textbf{100.0} & \textbf{97.0} & \textbf{98.7} & \textbf{82.0} & \textbf{76.2} & \textbf{83.5} & \textbf{81.0} & \textbf{89.0} \\ \hline
Admix \cite{32attackICCV2021admix}     & 74.6          & 96.2           & 87.7          & 92.7          & 55.7          & 52.3          & 52.3          & 50.0          & 70.2          \\
Admix+Ours & \textbf{85.9} & \textbf{96.9}           & \textbf{92.4} & \textbf{93.1} & \textbf{73.9} & \textbf{69.7} & \textbf{72.9} & \textbf{69.3} & \textbf{81.8} \\ \hline
SSA \cite{SSA2022ECCV}       & 88.2          & \textbf{100.0} & 95.8          & 97.5          & 68.2          & 67.5          & 72.9          & 69.0          & 82.4          \\
SSA+Ours   & \textbf{91.0} & \textbf{100.0} & \textbf{97.2} & \textbf{98.2} & \textbf{79.0} & \textbf{74.9} & \textbf{78.0} & \textbf{74.8} & \textbf{86.6} \\ \hline
\end{tabular}
\caption{The attack success rates (\%) of our method, when it is integrated with DIM, TIM, SIM, Admix, and SSA, respectively. The adversarial examples are generated on Res-50. Here * indicates the white-box model. The best results are bold.}
\label{tab:input}
\end{table*}

\begin{table*}[!htbp]
\centering
\footnotesize
\label{tab:ens}
\begin{tabular}{|l|cccccccc|c|}
\hline
Attack      & Inc-v3        & Res-50*        & Vgg-19*        & Den-121*       & ViT           & Swin          & Inc-v3$_{ens3}$    & Inc-v3$_{ens4}$    & Average       \\ \hline \hline
MI \cite{51CVPR2018mifgsm}         & 87.7          & 99.9           & 99.9           & 99.9           & 72.3          & 75.6          & 69.7          & 68.5          & 84.2          \\
VMI \cite{34CVPR2021vmifgsm}        & 93.9          & \textbf{100.0} & \textbf{100.0} & \textbf{100.0} & 85.6          & 86.8          & 85.6          & 82.4          & 91.8          \\
GRA \cite{GRAadaptive}        & 97.5          & \textbf{100.0} & \textbf{100.0} & \textbf{100.0} & 92.9          & 91.9          & 90.8          & 90.1          & 95.4          \\
PGN \cite{PGN_2023_NIPS}        & \textbf{97.6} & \textbf{100.0} & \textbf{100.0} & \textbf{100.0} & 93.3          & 92.6          & \textbf{92.4} & 90.1          & 95.8          \\
AdaMSI \cite{adamsi}     & 91.9          & \textbf{100.0} & \textbf{100.0} & \textbf{100.0} & 76.3          & 74.6          & 65.1          & 59.6          & 83.4          \\
TPA \cite{2024TPA}        & 95.3          & 98.7           & 98.7           & 98.8           & 89.7          & 90.6          & 90.1          & 88.3          & 93.8          \\
ResPA (Ours) & \textbf{97.6} & \textbf{100.0} & \textbf{100.0} & \textbf{100.0} & \textbf{94.0} & \textbf{93.2} & 92.3         & \textbf{90.5} & \textbf{96.0} \\ \hline
\end{tabular}
\caption{ The attack success rates (\%) on eight models under ensemble model setting. The adversarial examples are generated on  Res-50,  Vgg-19 and Den-121 models. Here * indicates the white-box model. The best results are bold.}
\end{table*}

\noindent\textbf{Parameter Setting.}
We set the maximum perturbation of the parameter $\epsilon = 16$, the step size $\alpha = 1.6$, and the number of iterations $T = 10$.
The decay factor $ \mu=1$ is set for all the approaches.
To ensure a fair comparison in this paper, we have adopted a uniform configuration for VMI, GRA, PGN, and TPA, specifying the number of sampled examples as $N = 5$ and defining the upper limit of the neighborhood size as $\beta = 1.5 \times \epsilon$.
For DIM, we set the transformation probability at 0.5. Regarding TIM, a Gaussian kernel of size $7 \times 7 $ is employed as in \cite{tifgsm}. In the case of SIM, the number of scale copies is set to $m = 5$. For Admix, we set the mixing ratio to 0.2 and the number of copies to 5.
For the proposed ResPA, we set the number of examples $N = 5$, the exponential decay factor $\theta = 0.6$, the balanced coefficient $\gamma = 0.6$, and the upper bound of $\beta = 1.5\times\epsilon$.


\subsection{Evaluation on Single Model}
In this section, we carry out multiple attacks.
The adversarial samples are crafted from three distinct
models, namely Inc-v3, Res-50, and Den-121, respectively.
The results are summarized in \cref{tab:black}. The models we attack are arranged in rows, and the eight models we test are arranged in columns.
From the results, it can be seen that the ResPA method proposed in this paper not only maintains a high attack performance in a white-box setting but also significantly improves the attack performance in a black-box setting.
For instance, when generating adversarial examples on Inc-v3, the state-of-the-art methods, namely GRA \cite{GRAadaptive}, PGN \cite{PGN_2023_NIPS}, and TAP \cite{2024TPA}, achieve average attack success rates of 59.6\%, 63.6\%, and 60.3\% on eight models, respectively. By contrast, ResPA attains an impressive average attack success rate of 64.6\%, outperforming them by 5.0\%, 1.0\%, and 4.3\% respectively.
Excellent results highlight that using the residual gradient as the perturbation direction can further enhance the attack performance of adversarial samples.

 \begin{figure*}[]
  \centering
  \subfloat[The hyper-parameter $\beta$]{\includegraphics[width=0.24\textwidth]{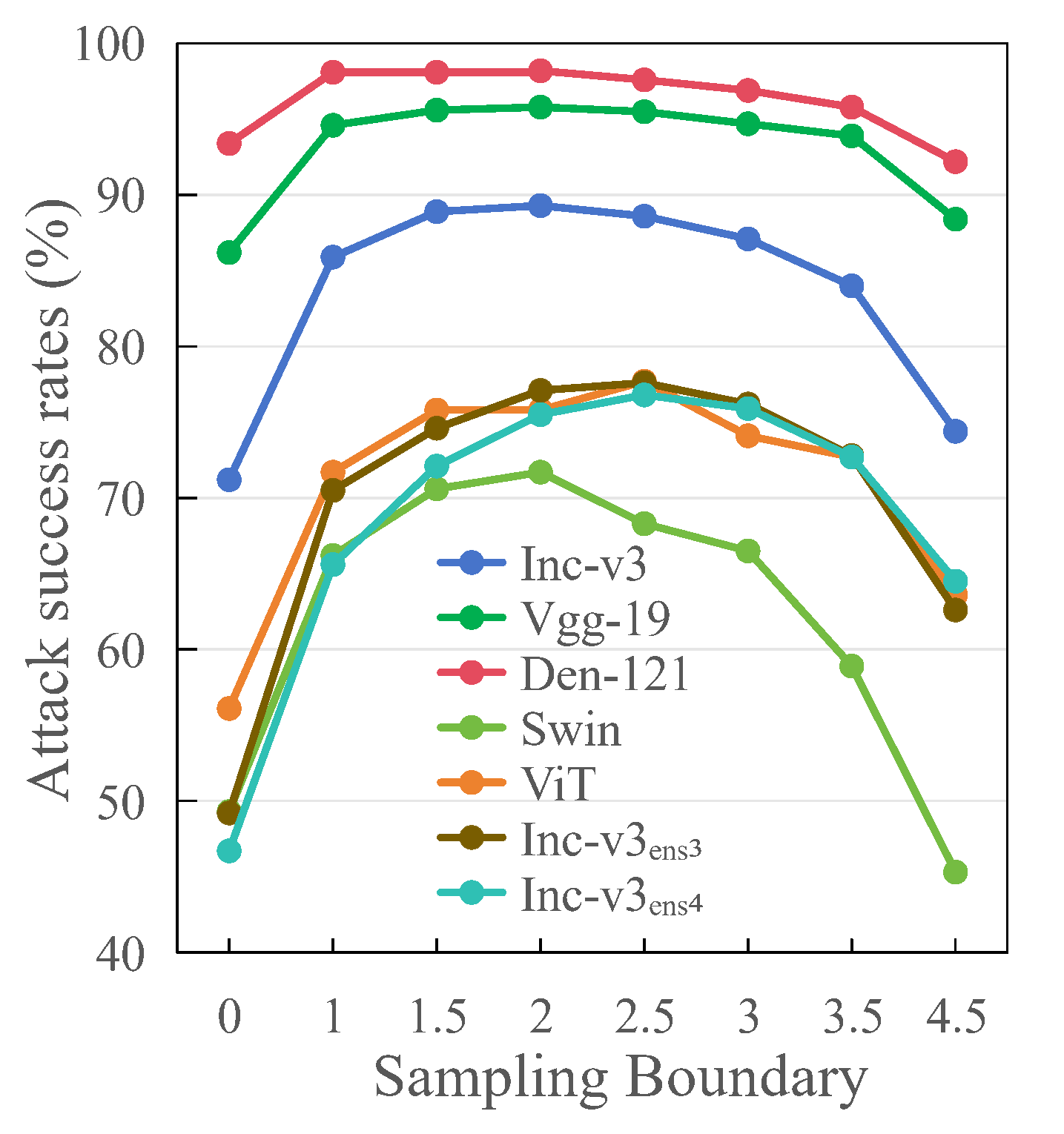}}
  \subfloat[The hyper-parameter $N$]{\includegraphics[width=0.235\textwidth]{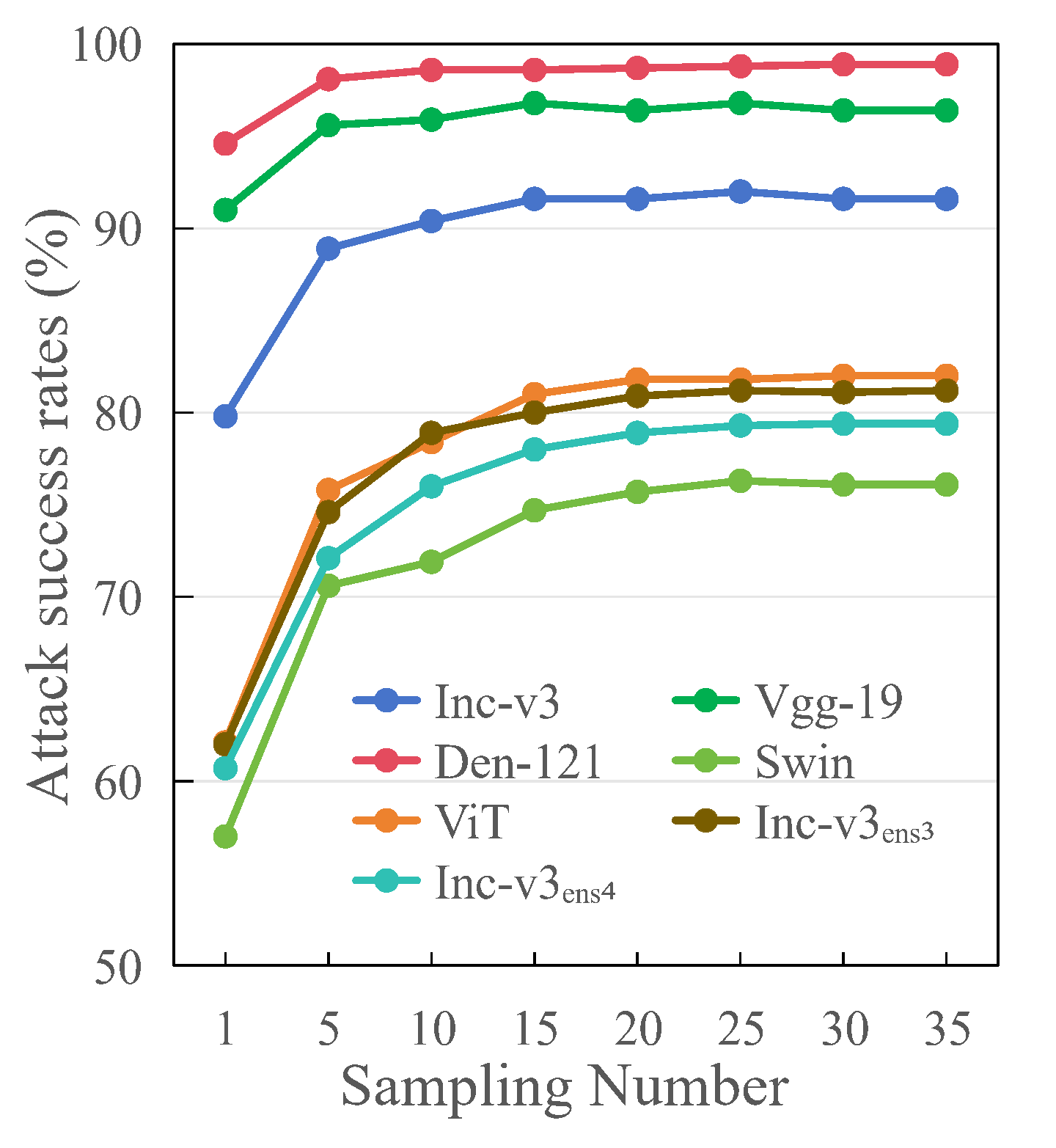}}
  \subfloat[The hyper-parameter $\theta$]{\includegraphics[width=0.24\textwidth]{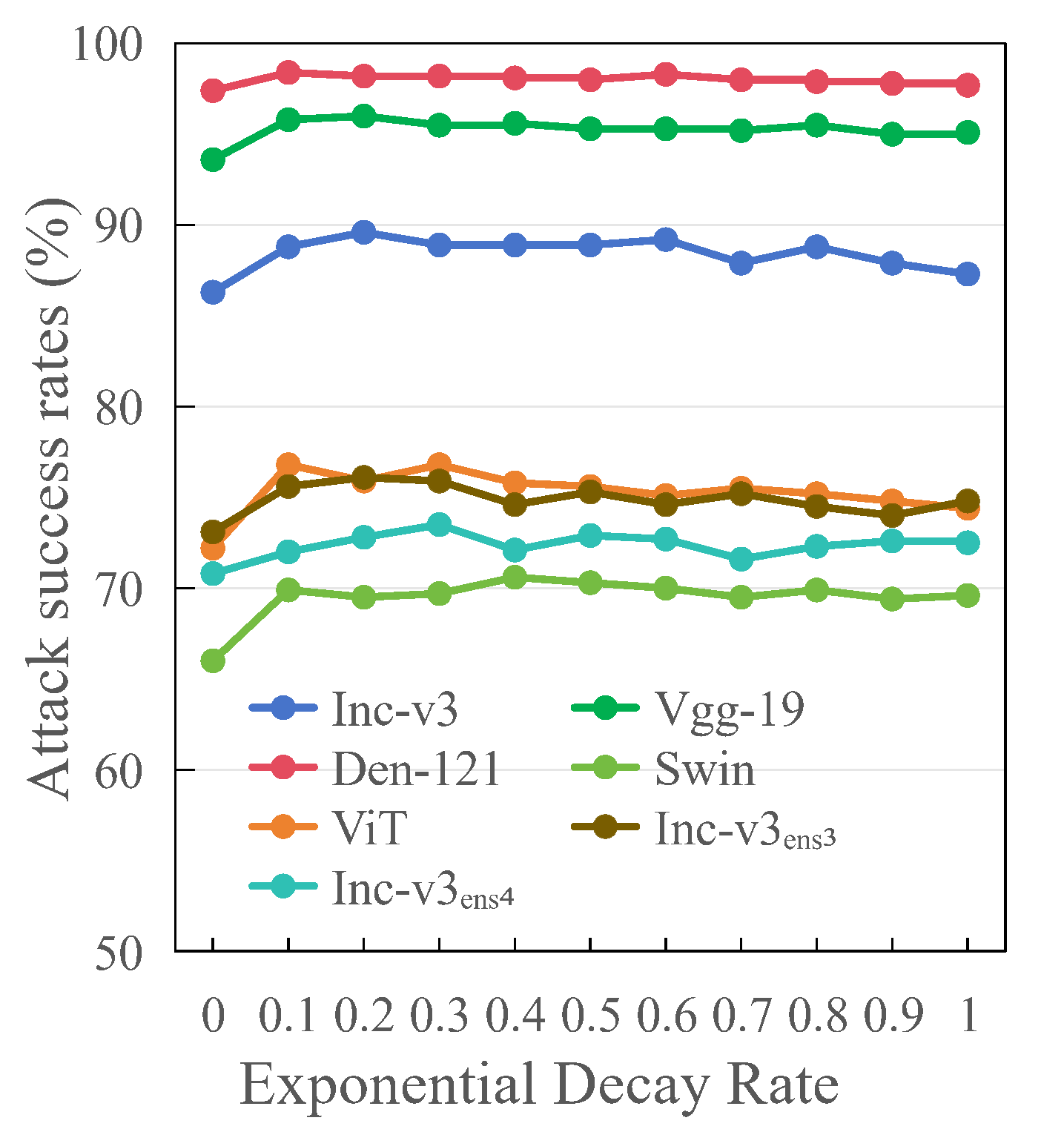}}
  \subfloat[The hyper-parameter $\gamma$]{\includegraphics[width=0.24\textwidth]{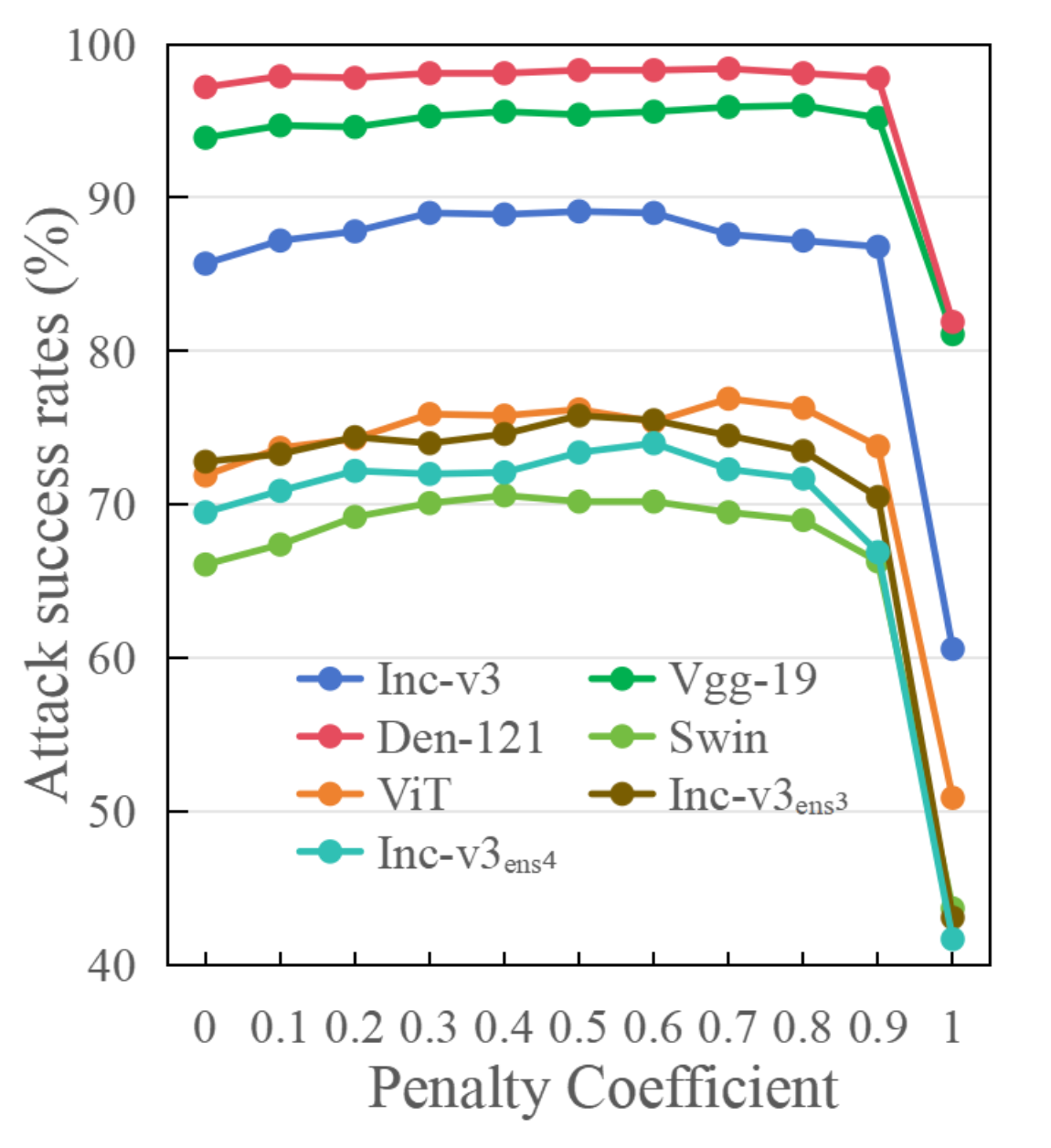}}
  \caption{The attack success rate (\%) on seven black-box models with different hyper-parameters $\beta$, $N$, $\theta$ and $\gamma$. The adversarial examples are generated by ResPA on Res-50.}
  \label{fig:study}
\end{figure*}

\subsection{Evaluation on Combined Input Transformation}

The attack success rates of combination approaches are also evaluated. Given the streamlined and effective gradient updating mechanism, ResPA can seamlessly integrate with input augmentation approaches to further enhance adversarial transferability.
We incorporate ResPA into five input augmentation attacks, namely DIM \cite{28attackDMICVPR2019}, TIM \cite{tifgsm}, SIM \cite{29attack2019NIsi}, Admix \cite{32attackICCV2021admix}, and SSA \cite{SSA2022ECCV}. All the combined methods create adversarial samples on Res-50, and the performances are presented in \cref{tab:input}. As can be seen from the table, the combinational attacks exhibit clear improvements over all baseline attacks. For instance, ResPA increases the average attack success rate of the five baseline attacks by 11.4\%, 20.3\%, 11.7\%, 11.6\%, and 4.2\%, respectively, which validates that ResPA can significantly enhance transferability.

In particular, after integrating these input transformation-based methods, ResPA is more likely to attain significantly superior outcomes on adversarially trained ensemble models in comparison with the results presented in \cref{tab:black}.
The adversarial samples created by the proposed ResPA method are situated within broader and smoother flat local maxima, which validates that the proposed method has the ability to generate adversarial examples located at the flat maximum.

\subsection{Evaluation on Ensemble Model}
We also evaluate the performance of ResPA in an ensemble-model setting.
The ensemble attack methodology described in \cite{51CVPR2018mifgsm} is adopted, constructing an ensemble by averaging the logit outputs from a diverse set of models.
Specifically, adversarial examples are crafted by integrating the predictions from three conventionally trained models: Res-50, Vgg-19, and Den-121.
Equal weights are assigned to all the ensemble models. Subsequently, we evaluate the transferability of standardly trained models and adversarially trained models and present the relevant results in \cref{tab:ens}. As can be seen from the table, compared with previous attacks, ResPA achieves the best performance. Importantly, when aimed at transformer-based models, ResPA constantly surpasses other transfer-based attacks. Furthermore, in the white-box setting, the proposed ResPA can still retain success rates similar to those of the baselines.

\begin{table*}[!htbp]
\centering
\small
\begin{tabular}{|l|ccccccc|c|}
\hline
Attack     & HGD \cite{HGD}           & Bit-Red \cite{bitred}      & FD \cite{FD2019feature}           & JPEG \cite{defenseJPEG}          & NRP \cite{NRP}          & R\&P \cite{RP2018mitigating}         & RS \cite{RS2019certified}           & Average       \\ \hline \hline
MI \cite{51CVPR2018mifgsm}        & 40.2          & 33.2          & 44.2          & 36.2          & 32.6          & 42.2          & 28.7          & 36.8          \\
VMI \cite{34CVPR2021vmifgsm}       & 60.0          & 51.2          & 61.8          & 56.0          & 42.3          & 59.7          & 33.0          & 52.0          \\
GRA \cite{GRAadaptive}       & 64.0          & 54.1          & 64.9          & 59.2          & 43.2          & 62.4          & 35.8          & 54.8          \\
PGN \cite{PGN_2023_NIPS}       & 68.5          & 59.8 & 69.0          & 64.8          & 45.8          & 66.9          & 36.9          & 58.8          \\
AdaMSI \cite{adamsi}     & 38.9          & 34.0          & 43.5          & 37.3          & 22.4          & 41.9          & 29.9          & 35.4          \\
TPA \cite{2024TPA}       & 63.8          & 58.8          & 66.9          & 61.0          & 44.9          & 62.8          & 37.4 & 56.5          \\
ResPA (Ours) & \textbf{69.2} & \textbf{60.7} & \textbf{69.9} & \textbf{65.2} & \textbf{48.0} & \textbf{67.9} & \textbf{37.7} & \textbf{59.8} \\ \hline
\end{tabular}
\caption{The attack success rates (\%) of seven advanced defense mechanisms on adversarial samples. The adversarial samples are generated on the Inc-v3 model by various transfer-based attacks. The best results are bold.}
\label{tab:defense}
\end{table*}

\begin{figure*}[]
\centering
\includegraphics[width=0.75\textwidth]{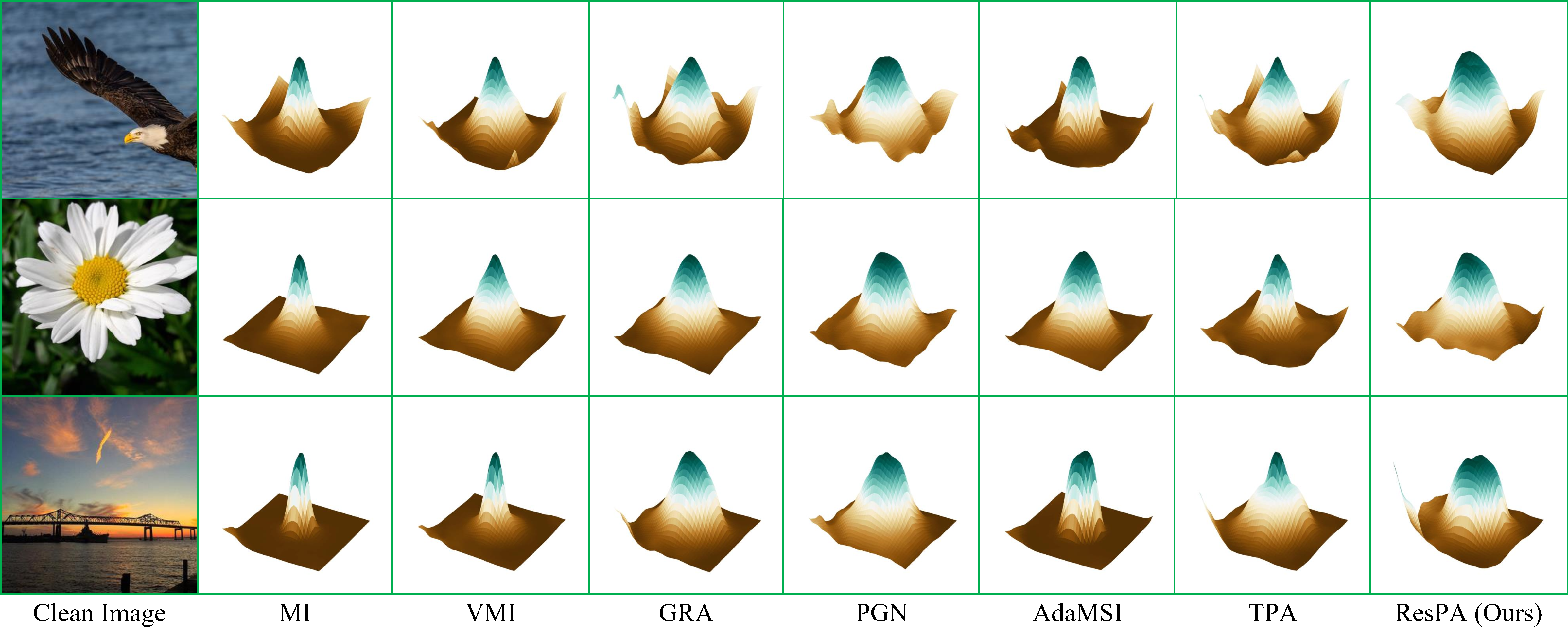}
\caption{Visualization of loss surfaces along two random directions for three randomly sampled
adversarial examples on the surrogate model (Inc-v3). Compared with other methods, ResPA can assist adversarial examples in reaching flatter maximum regions.}
\label{fig:loss}
\end{figure*}

\subsection{Evaluation on Defense Method}
To evaluate the effectiveness of the proposed ResPA method, we also examine the attack success rates of ResPA against various advanced defense mechanisms.
Some advanced defense methods, such as HGD, Bit-Red, FD, JPEG, NRP, R\&P, and RS, are employed.
Results are presented in  \cref{tab:defense}. In the black-box setting, it is observed that ResPA outperforms other state-of-the-art attack algorithms.
For example, GRA \cite{GRAadaptive}, TPA \cite{2024TPA}, and PGN \cite{PGN_2023_NIPS}  attain average success rates of 54.8\%, 56.5\%, and 58.8\%, respectively, when tested against the seven defense models. In comparison, the proposed ResPA approach attains an average success rate of 59.8\%, outperforming them by 4.8\%, 3.3\%, and 1.0\%, respectively.
This significant progress fully demonstrates the outstanding effectiveness of ResPA when dealing with adversarially trained models and other defense models.

\subsection{Attacking Person Re-identification}
We also conduct comparative experiments on the person re-identification (Re-ID) benchmark dataset \cite{10datasetmarket}. To successfully attack the Re-ID system, we use predicted labels instead of ground truth labels. The queries of the Re-ID system are attacked as adversarial queries, targeting different backbone networks of the Re-ID model, including DenseNet-121 (Den-121) \cite{densenet}, ConvNext (Conv) \cite{49convnet}, Swin-Transformer (Swin) \cite{liu2021Swin}, and Swinv2-Transformer (Swinv2) \cite{48swinv2}.
The evaluation metrics are Rank-1 and mAP, where lower values indicate better attack performance. As shown in \cref{tab:market}, the results demonstrate that the proposed ResPA achieves the best attack performance compared to state-of-the-art methods.

\begin{table*}[]
\centering
\footnotesize
\begin{tabular}{|l|llllllllll|}
\hline
\multicolumn{1}{|c|}{\multirow{3}{*}{Method}} & \multicolumn{10}{c|}{Market-1501}                                                                                                                                                     \\ \cline{2-11}
\multicolumn{1}{|c|}{}                        & \multicolumn{2}{c|}{Den-121}                                     & \multicolumn{2}{c|}{Conv}                                        & \multicolumn{2}{c|}{Swin}                                        & \multicolumn{2}{c|}{Swinv2}                                      & \multicolumn{2}{c|}{Average}                          \\ \cline{2-11}
\multicolumn{1}{|c|}{}                        & \multicolumn{1}{c}{Rank-1} & \multicolumn{1}{c|}{mAP}            & \multicolumn{1}{c}{Rank-1} & \multicolumn{1}{c|}{mAP}            & \multicolumn{1}{c}{Rank-1} & \multicolumn{1}{c|}{mAP}            & \multicolumn{1}{c}{Rank-1} & \multicolumn{1}{c|}{mAP}            & \multicolumn{1}{c}{Rank-1} & \multicolumn{1}{c|}{mAP} \\ \hline  \hline
Before attack                                 & 89.22                      & \multicolumn{1}{l|}{73.62}          & 89.88                      & \multicolumn{1}{l|}{72.57}          & 92.43                      & \multicolumn{1}{l|}{78.90}          & 91.54                      & \multicolumn{1}{l|}{77.67}          & 90.77                      & 75.69                    \\
MI  \cite{51CVPR2018mifgsm}                                          & 32.01                      & \multicolumn{1}{l|}{22.20}          & 43.38                      & \multicolumn{1}{l|}{29.39}          & 54.66                      & \multicolumn{1}{l|}{40.95}          & 51.22                      & \multicolumn{1}{l|}{38.11}          & 45.32                      & 32.66                    \\
VMI  \cite{34CVPR2021vmifgsm}                                         & 20.37                      & \multicolumn{1}{l|}{14.70}          & 29.25                      & \multicolumn{1}{l|}{19.61}          & 41.51                      & \multicolumn{1}{l|}{30.63}          & 37.20                      & \multicolumn{1}{l|}{27.41}          & 32.08                      & 23.09                    \\
PGN  \cite{PGN_2023_NIPS}                                         & 18.91                      & \multicolumn{1}{l|}{13.43}          & 26.37                      & \multicolumn{1}{l|}{17.84}          & 39.76                      & \multicolumn{1}{l|}{28.85}          & 35.93                      & \multicolumn{1}{l|}{26.14}          & 30.24                      & 21.57                    \\
GRA  \cite{GRAadaptive}                                         & 21.85                      & \multicolumn{1}{l|}{15.44}          & 31.86                      & \multicolumn{1}{l|}{21.30}          & 43.91                      & \multicolumn{1}{l|}{32.49}          & 39.73                      & \multicolumn{1}{l|}{29.44}          & 28.88                      & 20.81                    \\
BSR  \cite{2024BSR}                                         & 18.62                      & \multicolumn{1}{l|}{13.16}          & 27.49                      & \multicolumn{1}{l|}{18.39}          & 40.71                      & \multicolumn{1}{l|}{29.90}          & 37.74                      & \multicolumn{1}{l|}{27.04}          & 31.14                      & 22.12                    \\
TPA \cite{2024TPA}                                          & 15.47                      & \multicolumn{1}{l|}{11.30}          & 24.26                      & \multicolumn{1}{l|}{16.09}          & 35.78                      & \multicolumn{1}{l|}{26.12}          & 32.60                      & \multicolumn{1}{l|}{23.30}          & 27.03                      & 19.20                    \\
ResPA (Ours)                                  & \textbf{15.30}             & \multicolumn{1}{l|}{\textbf{11.12}} & \textbf{24.14}             & \multicolumn{1}{l|}{\textbf{15.95}} & \textbf{35.31}             & \multicolumn{1}{l|}{\textbf{25.67}} & \textbf{31.03}             & \multicolumn{1}{l|}{\textbf{22.33}} & \textbf{26.45}             & \textbf{18.77}           \\ \hline
\end{tabular}
\caption{Performance (\%) of adversarial attacks against the four Re-ID models under black-box setting on the Market-1501
dataset. The adversarial queries are crafted on Res-50. Lower is better for the attack.}
\label{tab:market}
\end{table*}

 \subsection{Experiments on Hyper-parameters}
 Various experiments are conducted about the hyper-parameters of ResPA, namely the sampling boundary $\beta$, the sampling number $N$, the exponential decay rate $\theta$, and the penalty coefficient $\gamma$. For the sake of simplified analysis, all adversarial examples are generated based on the Res-50 model. By default, we set $\beta = 1.5\times\epsilon$, $N = 5$, $\theta = 0.4$, and $\gamma = 0.4$.

\noindent\textbf{The sampling boundary $\beta$.}
We analyze the influence of the sampling boundary $\beta$ on the result. As shown in  \cref{fig:study}(a), as $\beta$ increases, the transferability improves, and when $\beta = 1.5\times\epsilon$, it reaches the peak for CNN-based models. However, when facing Transformer-based and adversarially trained models, the transferability still increases. When $\beta>2.5\times\epsilon$, the performance of adversarial transferability will decline on seven black-box models. For a fair comparison, we uniformly set $\beta = 1.5\times\epsilon$ for the methods \cite{34CVPR2021vmifgsm,GRAadaptive,PGN_2023_NIPS,2024TPA} involving sampling.

\noindent\textbf{The sampling number $N$.}
In  \cref{fig:study}(b), we explore the influence of $N$. As the $N$ increases, the transferability also rises. In the case of CNN-based models, when $N>15$,  the performance of adversarial transferability gradually stabilizes on 7 black-box models. Nevertheless, for Transformer-based and adversarially trained models, the transferability continues to increase. When $N>25$, the attack success rates gradually stabilize across seven black-box models.  For a fair comparison, we uniformly set $N=5$ for the methods \cite{34CVPR2021vmifgsm,GRAadaptive,PGN_2023_NIPS,2024TPA} involving sampling.

\noindent\textbf{The exponential decay rate $\theta$.}
We investigate the influence of $\theta$ on the results. As shown in  \cref{fig:study}(c), as $\theta$ increases, the transferability improves. When $\theta > 0.2$, the transferability of these black-box models is almost stable. This also indicates that when $\theta$ is in the interval [0.2, 1], it has little influence on the transferability. In this paper, $\theta$ is set to 0.6.

\noindent\textbf{The penalty coefficient $\gamma$.}
As depicted in  \cref{fig:study}(d), we examine the influence of $\gamma$. In particular, when $\gamma$ lies within the interval $[0.2, 0.9]$, the transferability of these black-box models is relatively satisfactory. However, when $\gamma > 0.9$, the transferability declines sharply. In this paper, $\gamma=0.6$.


\subsection{Visualization of Loss Surfaces}
To verify that ResPA can help adversarial examples find a flat maxima area, we compare the loss surface maps of adversarial examples generated by different attacking approaches on the Inc-v3 model.
Each 2D graph corresponds to an adversarial sample, with the adversarial sample placed at the center. Each row in  \cref{fig:loss} represents the visualization of one image. Compared with other methods, ResPA can assist adversarial examples in reaching flatter maxima areas. The adversarial examples created by ResPA are located in broader and smoother flat areas, which validates that ResPA can create adversarial samples located in the flat maximum.



\section{Conclusion}


In this paper, we propose ResPA, a novel attack method to identify highly transferable adversarial examples. Instead of directly focusing on the current gradient as the perturbation direction, ResPA considers the residual between the current and historical gradient as the perturbation direction, thereby trying to avoid the over-reliance on the perturbed point in excessively sharp regions. As a byproduct, ResPA incorporates the proposed flatness term as a regularization to maximize the loss function while making the loss surface flatter. Experimental results demonstrate the better transferability of ResPA than the existing state-of-the-art transfer-based attack approaches.

\section{Acknowledgement}
This research is supported by National Natural Science Foundation of China (U21A20470, 62172136, 72188101); Institute of Advanced Medicine and Frontier Technology (2023IHM01080); Liaoning Provincial Natural Science Foundation (2024-MS-012); National Key Research and Development Program of china (2024YFB4710800); Dalian Science and Technology Talent Innovation Support Plan (2024RY010); Natural Science Foundation of Hebei Province (F2025201037).



{
    \small
    \bibliographystyle{ieeenat_fullname}
    \bibliography{main}
}

\end{document}